\pdfoutput=1

\documentclass[11pt]{article}

\usepackage[]{acl}

\usepackage{times}
\usepackage{latexsym}
\usepackage[prependcaption,textsize=tiny]{todonotes}

\usepackage[T1]{fontenc}

\usepackage[utf8]{inputenc}
\usepackage{adjustbox}
\usepackage{multirow}
\usepackage[normalem]{ulem}
\usepackage{wrapfig}
 \usepackage{booktabs}
 \usepackage{subfig}
\usepackage{microtype}
\usepackage{soul}

\title{A Novel Multi-Task Learning Approach for  Context-Sensitive Compound Type Identification in Sanskrit}

\newcommand*{\affmark}[1][*]{\textsuperscript{#1}}

\author{Jivnesh Sandhan\affmark[1], Ashish Gupta\affmark[2], Hrishikesh Terdalkar\affmark[1], Tushar Sandhan\affmark[1],\\\textbf{Suvendu Samanta\affmark[1], Laxmidhar Behera\affmark[1,3] and Pawan Goyal\affmark[2]}\\
\affmark[1]IIT Kanpur, \affmark[2]IIT Kharagpur, \affmark[3]IIT Mandi\\
\texttt{jivnesh@iitk.ac.in, ashishgupta2598@gmail.com,}\\
\texttt{pawang@cse.iitkgp.ac.in}}

\begin{document}
\maketitle

\begin{abstract}
 The phenomenon of compounding is ubiquitous in Sanskrit. It serves for achieving brevity in expressing  thoughts, while simultaneously enriching the lexical and structural formation of the language. In this work, we focus on the {\sl Sanskrit Compound Type Identification (SaCTI)} task, where we consider the problem of identifying semantic relations between the components of a compound word. Earlier approaches solely rely on the lexical information obtained from the components and ignore the most crucial contextual and syntactic information useful for SaCTI. However, the SaCTI task is challenging primarily due to the implicitly encoded context-sensitive semantic relation between the compound components.

Thus, we propose a novel multi-task learning architecture which incorporates the contextual information and enriches the complementary syntactic information using morphological tagging and dependency parsing as two auxiliary tasks. Experiments on the benchmark datasets for SaCTI show $6.1$ points (Accuracy) and $7.7$ points (F1-score) absolute gain compared to the state-of-the-art system. Further, our multi-lingual experiments demonstrate the efficacy of the proposed architecture in English and Marathi languages.\footnote{The code and datasets are publicly available at: \url{https://github.com/ashishgupta2598/SaCTI}}
\end{abstract}

\section{Introduction}
A compound is defined as a collection of one or more entities that act as a single meaningful entity. The process of decoding an implicit semantic relation between the components of a compound in Sanskrit is called {\sl Sanskrit Compound Type Identification (SaCTI)}. Alternatively, it is also termed as {\sl Noun Compound Interpretation (NCI)} \cite{ponkiya-etal-2021-framenet,ponkiya-etal-2020-looking}. In the literature, the NCI problem has been formulated in two related but different ways. Let's take {\sl mango juice} as an example. In the first formulation, the relation between the two components is labeled from  a set of semantic relations ({\sl MADEOF}) \cite{dima-hinrichs-2015-automatic,fares-etal-2018-transfer,ponkiya-etal-2021-framenet}. The second formulation uses paraphrasing to illustrate the semantic relations ({\sl a juice made from mango}) \cite{lapata-keller-2004-web,ponkiya-etal-2018-treat,ponkiya-etal-2020-looking}.
 
In this work, we use the first formulation that frames the task as a multi-class classification problem.
The task is challenging and often depends upon the context or world knowledge about the entities involved \cite{krishna-etal-2016-compound}. For instance, the semantic type of the compound {\sl r\={a}ma-\={\i}\'{s}vara\d{h}}  can be classified into one of the following semantic types depending on the context: {\sl Karmadh\={a}raya}\footnote{There are 4 broad semantic types of compounds: \textit{Avyay\={\i}bh\={a}va}, \textit{Bahuvr\={\i}hi}, \textit{Dvandva}, and \textit{Tatpuru\d{s}a}. {\sl Karmadh\={a}raya} is considered as sub-type of \textit{Tatpuru\d{s}a}. We encourage readers to refer \newcite{krishna-etal-2016-compound} for more details on these semantic types.}, {\sl Bahuvr\={\i}hi}  and {\sl Tatpuru\d{s}a}.
Although the compound has the same components as well as the final form, the implicit relationship between the components can be decoded only with the help of available contextual information \cite{kulkarni2013,krishna-etal-2016-compound}. Due to such instances, the downstream Natural Language Processing (NLP) applications for Sanskrit such as question answering \cite{terdalkar-bhattacharya-2019-framework} and machine translation \cite{aralikatte-etal-2021-itihasa}, etc. show sub-optimal performance when they stumble on compounds. 
For example, while translating {\sl r\={a}ma-\={\i}\'{s}vara\d{h}} into English, depending on the semantic type, there are three possible meanings: (1) Lord who is pleasing (in {\sl Karmadh\={a}raya})
(2) the one whose God is Rama (in {\sl Bahuvr\={\i}hi}) (3) Lord of Rama (in {\sl Tatpuru\d{s}a}).
Therefore, the SaCTI task can be seen as a preliminary pre-requisite to building a robust NLP technology for Sanskrit. Further, this dependency on contextual information rules out the possibility of storing and doing a lookup to identify a compound's semantic types. 

 With the advent of recent contextual models \cite{peters-etal-2018-deep,devlin-etal-2019-bert,conneau-etal-2020-unsupervised}, there has been upsurge in performance of various downstream NLP applications \cite{kondratyuk-straka-2019-75,roberta,NEURIPS2019_dc6a7e65}.
 Nevertheless, there have been no efforts to build context-dependent models in SaCTI.\footnote{Refer to related work section (\S~\ref{related_section}) for more details.}  This may be attributed to the fact that while most of the natural language technology is built for resource-rich languages such as English \cite{joshi-etal-2020-state}, compounding is not a predominant phenomenon in them \cite{krishna-etal-2016-compound}. 
 There is also lack of task-specific context-sensitive labeled data.

Earlier approaches \cite{kulkarni2013,krishna-etal-2016-compound,sandhan-etal-2019-revisiting} for SaCTI solely rely on lexical information obtained from the components and are blind to potentially useful contextual and syntactic information. 
The context is the most feasible, cheaply available information. 
As per Pa\d{n}ini's grammar \cite{panini,kulkarni2013}, the morphological features have direct correlation with the semantic types. Sometimes, the dependency information also helps is disambiguation and can provide a medium to enrich contextual information. Thus, we propose a novel multi-task learning approach which (1) incorporates the contextual information, and (2) enriches the complementary syntactic information using morphological tagging and dependency parsing auxiliary tasks without any additional manual labeling. 
  Summarily, our key contributions are:
  \begin{itemize}
      \item We propose a novel context-sensitive multi-task learning architecture for SaCTI (\S~\ref{proposed-system}).
      \item We illustrate that morphological tagging and dependency parsing auxiliary tasks  are helpful and serve as a proxy for explainability of system predictions (\S~\ref{analysis}) for the SaCTI task.
      \item We report results with $7.71$ points (F1) absolute gains compared to the current state-of-the-art system by \newcite{krishna-etal-2016-compound} (\S~\ref{main-results}). 
       \item We show the efficacy of the proposed approach in English and Marathi (\S~\ref{analysis}).
       \item We release our codebase and pre-processed datasets (including newly annotated Marathi dataset (\S~\ref{datasets})) and web-based tool (\S~\ref{analysis}) for using our pretrained models.  
  \end{itemize}

\section{The proposed system}
\label{proposed-system}
\begin{figure}[!htb]
\centering
\includegraphics[width=0.45\textwidth]{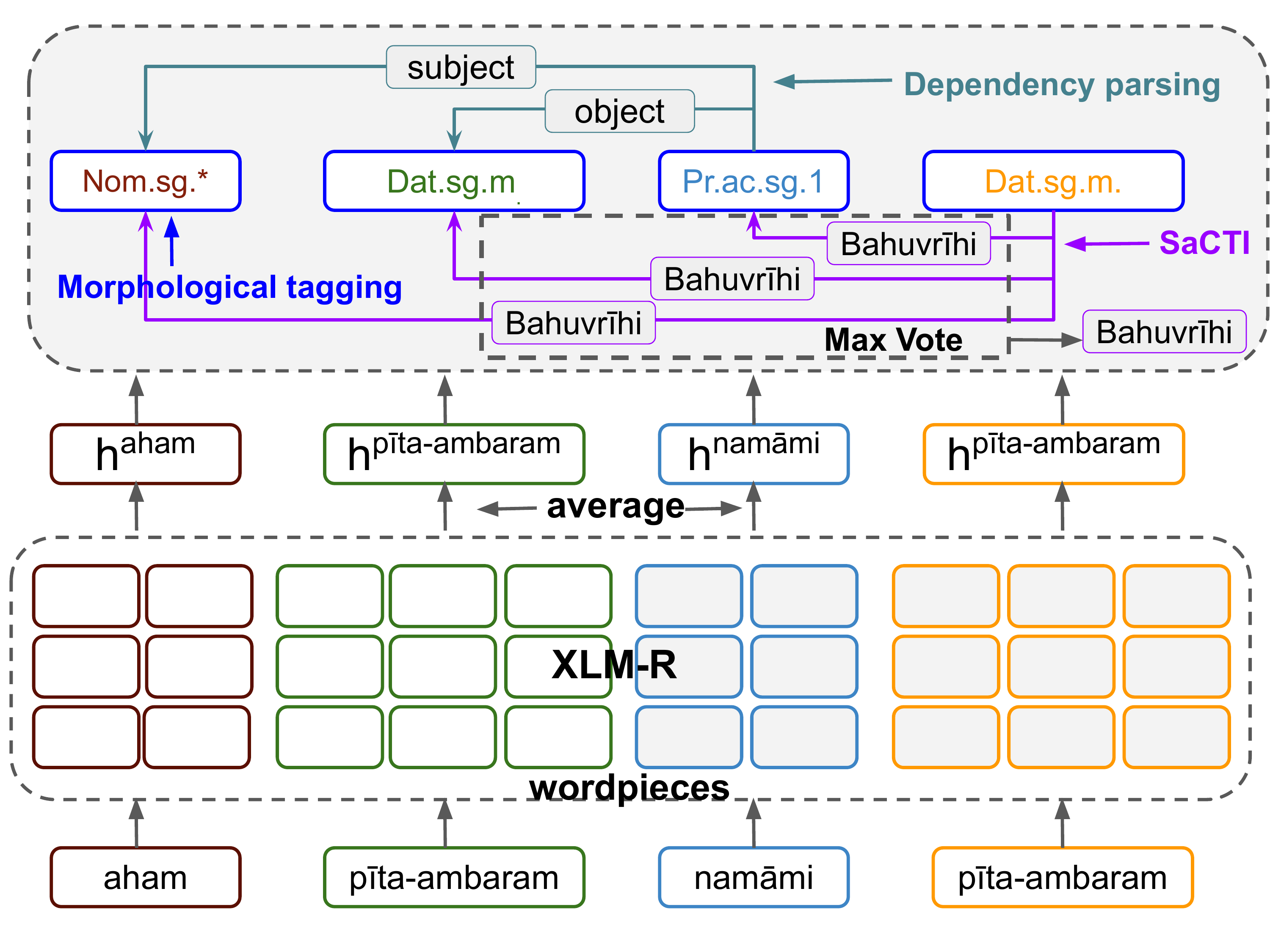}
\caption{Illustration of the proposed multi-task learning architecture with an example ``{\sl aham p\={\i}ta-ambaram nam\={a}mi}'' (Translation: ``I pray to P\={\i}t\={a}mbara (Lord Vishnu).'') where `{\sl p\={\i}ta-ambaram}' is a compound word belonging to the {\sl Bahuvr\={\i}hi} semantic type as per the given context.  We feed this context and the compound word at the end as an input to the system. Each token is split into wordpieces using a multi-lingual tokenizer \cite{kudo-richardson-2018-sentencepiece}. This sequence of wordpieces is passed to multi-lingual pretrained XLM-R encoder \cite{conneau-etal-2020-unsupervised}. The hidden representation of each token is the average of its wordpieces' representations obtained from the encoder. We apply our multi-task learning architecture which consists of three tasks, namely, Sanskrit compound type identification (SaCTI), morphological tagging and dependency parsing over the hidden representations. We formulate SaCTI as a pair-wise (a context word and the compound) classification task where the objective is to predict the semantic type of the target compound word ({\sl Bahuvr\={\i}hi}). At test time, we apply the maximum voting policy to select a single prediction from the set of semantic relations predicted by such $n$ pairs.} 
\label{fig:main_model} 
\end{figure}
Figure~\ref{fig:main_model} illustrates the proposed multi-task learning architecture with an example context, ``{\sl aham p\={\i}ta-ambaram nam\={a}mi}'' (Translation: ``I pray to P\={\i}t\={a}mbara (Lord Vishnu).'') where `{\sl p\={\i}ta-ambaram}' is a compound belonging to the {\sl Bahuvr\={\i}hi} semantic type as per the given context.  As shown in Figure~\ref{fig:main_model}, we feed this context along with the compound word concatenated at the end, as an input to the system, and obtain hidden representations from the multi-lingual encoder as described below.  
On top of the hidden representations as obtained via the encoder module, we apply our multi-task learning architecture consisting of three tasks: SaCTI, morphological tagging, and dependency parsing. We formulate the SaCTI task as a pair-wise (a context word paired with the compound word) classification task where the objective is to predict the semantic type of the target compound word ({\sl Bahuvr\={\i}hi} for {\sl p\={\i}ta-ambaram} compound word in this example) for all the pairs as shown in Figure~\ref{fig:main_model}. At test time, we apply the maximum voting policy 
to select a single prediction from the set of semantic types predicted by these pairs. We formally discuss the details of each component below.

\paragraph{Multilingual Encoder:} Sanskrit is a morphologically rich and low-resource language. In order to build powerful contextual representations for Sanskrit words, morphological richness poses the out-of-vocabulary problem and low-resource nature poses an unlabelled data scarcity problem. Thus, we opt for a multi-lingual encoder \cite[XLM-R]{conneau-etal-2020-unsupervised} to mitigate these issues. 

Given a compound $c_p$ and its context $C= [c_1, c_2, ..., c_n]$ such that $p^{th}$ position $(1 \leq p \leq n)$ in the context is the compound word, we append the compound word to the context such that $C= [c_1, c_2, ..., c_n, c_{n+1}]$ where $c_{n+1} = c_p$. Each token $(c_i)$ is further split into wordpieces $(c_i = [c_i^1,c_i^2,..., c_i^{m_i}])$ using a multi-lingual subword tokenizer \cite[Sentencepiece]{kudo-richardson-2018-sentencepiece,kudo-2018-subword}. Next, we pass the overall sequence $(C = [c_1^1,c_1^2,..., c_n^{m_n},c_{n+1}^{1}..., c_{n+1}^{m_{n+1}}])$ of wordpieces corresponding to the context $C$ into the pretrained transformer. Finally, we obtain the contextual representation of all tokens as $h = (h_1, h_2, h_3, ..., h_n, h_{n+1})$ where 
\begin{equation}
    h_i = \frac{1}{m_i}\sum_{k=1}^{m_i}Transformer(c_i^k)
\end{equation}
\paragraph{SaCTI:} Our context-sensitive classifier uses Bi-Affine attention, henceforth referred to as BiAFF. 
Given the hidden representations of $i^{th}$, $j^{th}$ context words as $h_i$, $h_j$ from the multi-lingual encoder, the scoring function $s_{i,j}$ 
indicates the system's belief that the
latter $(j^{th})$ (Eqn.~\ref{eq1}) should be related to the former $(i^{th})$  in identifying the semantic type of the latter, where $q_i^{T}z_i$ indicates bias to capture the prior of contextual information in the $i^{th}$ word.
\begin{equation}
s_{i,j} = z_i^{T}Uz_j + q_i^{T}z_i \\ \label{eq1}
\end{equation}
where $z_i = MLP(h_i)$, $U$ and $q_i$ are learnable parameters, $MLP$ denotes a multi-layered perceptron. 
Similarly, a score for $k^{th}$ possible semantic type relation between every pair of $i^{th}$ context word ($\forall i \in [1,n]$) and the compound is computed by:
\begin{equation}
r_{i,k} = z_i^{'T}U_k^{'}z_{n+1}^{'} + q_k^{'T}[z_i^{'};z_{n+1}^{'}] +b_k^{'}\\
\end{equation}
where $z_i^{'} = MLP^{'}(h_i)$, $U^{'}$, $b_k^{'}$ and $q_k^{'}$ are learnable parameters. Finally, model maximizes the following objective function during training.
\begin{equation}
  \sum_{i=1}^{n} p(y_{n+1}|c_i,\theta) + p(y_{n+1}^l|c_i,y_{n+1},\theta) \\ \label{eq3}
\end{equation}
where $y_{n+1}$ is the target compound appended in context $C$, $y_{n+1}^{l}$ is the semantic type of $(y_{n+1}/c_i,\theta)$, $\theta$ denotes system's parameters, $p(y_{n+1}|c_i,\theta) \propto exp(s_{i,n+1})$, $p(y_{n+1}^l|c_i,y_{n+1},\theta) \propto exp(r_{i,l})$. At test time, we apply maximum voting policy to select a single prediction from the set of semantic relations predicted for $n$ pairs $(c_i,c_{n+1}), \forall i \in [1,n]$. 

\paragraph{Morphological tagging:}  The primary motivation behind using morphological tagging as an auxiliary task aligns well with grammatical rules from Pa\d{n}ini's grammar \cite{panini,kulkarni2013}. For instance, {\sl Avyay\={\i}bh\={a}va} compounds are in neuter gender. {\sl Tatpuru\d{s}a} is a function of `case' attribute of morphological features. The number attribute of a compound depends on the semantic type of the compound. Also, there are constraints based on inflection/derivational suffix.  Summarily, these morphological features have direct correlation with the semantic classes. 
In our proposed system, the morphological tagging task leverages hidden representations from the multi-lingual encoder and decodes the pseudo-labels\footnote{The benchmark datasets do not have a gold standard morphological information. We use predicted morphological information as pseudo-labels (\S~\ref{datasets}).} using a fully connected layer followed by a softmax layer.\footnote{Note that all the parameters present in the multi-lingual encoder are trainable during the task-specific training.} 
In this process, morphological information useful for the SaCTI task is enriched in the hidden representations. This can be seen as an implicit way to encode the grammatical rules in the system.

\paragraph{Dependency parsing:} 
\newcite{syntax} argues that compounding is mostly a syntactic phenomenon. {\sl Bahuvr\={\i}hi} compounds are ``exocentric'' in nature, which attribute a property to an entity external to the compound with the adjective relationship.  Sometimes, syntactic information can provide a complementary signal useful for compound type disambiguation. For instance, consider the following example: {\sl aham n{\=\i}la-utpala\d{h} sara\d{h} pa\'sy\=ami} (Translation: I watch the pond having a blue-lotus.) Here, {\sl n{\=\i}la-utpalam} qualifies to be {\sl Bahuvr\={\i}hi} due to presence of its referent {\sl sara\d{h}} with an adjective relationship. However, in the absence of {\sl sara\d{h}} in the context, ambiguity pops up in between {\sl Bahuvr\={\i}hi} and {\sl Tatpuru\d{s}a}.\footnote{The ambiguity is whether I am seeing the blue lotus or the pond having a blue lotus.}
This motivates us to investigate the usefulness of syntactic information for the SaCTI task. The benchmark datasets do not have a gold standard dependency information. We use predicted dependency trees as pseudo-labels (\S~\ref{datasets}).  Our dependency parsing component leverages Bi-Affine parser \cite{DBLP:conf/iclr/DozatM17} over hidden representations from multi-lingual encoder.

\section{Experiments}
\subsection{Dataset and metrics}
\label{datasets}
Table~\ref{table:data} reports the unique number of compounds, data statistics and the number of semantic types for the respective datasets used in this work.  We restrict to binary compounds (compounds with two components) in all the datasets. These datasets consist of components, context and semantic type of a compound. The SaCTI datasets for Sanskrit are available with two levels of annotations: coarse (4 broad types) and fine-grained (15 sub-types).
\begin{table}[h]
\centering
\begin{adjustbox}{width=0.45\textwidth}
\small
\begin{tabular}{|c|c|c|c|c|c|}
\hline
\textbf{Datasets}  &\textbf{\#Unique}  & \textbf{\#Train}     & \textbf{\#Dev}     & \textbf{\#Test}  & \textbf{\#Types}    \\ \hline
SaCTI-base   &8,594& 9,356  & 2,339 &2,994& 4 (15) \\\hline
SaCTI-large &48,132&59,133 & 6,571 & 7,301 & 4 (15)  \\\hline
English &4,676& 4,163 & 1,041 & 1,301 & 7 \\\hline
Marathi  &368&659 &99  &114  &4 \\\hline
\end{tabular}
\end{adjustbox}
\caption{Data statistics for all the datasets} 
\label{table:data}
\end{table}

\paragraph{Sanskrit:} We evaluate on two available context-sensitive benchmark datasets: {\sl SaCTI-base} and {\sl SaCTI-large}.
We follow the same experimental settings as \newcite{krishna-etal-2016-compound} in {\sl SaCTI-base} to keep our results comparable with their state-of-the-art results. {\sl SaCTI-base} is a subset of {\sl SaCTI-large} dataset. In due course of time, more annotated data resulted in {\sl SaCTI-large} dataset.

\paragraph{English:} We use instance-based (context-dependent) noun-noun compound dataset released by \newcite{fares-2016-dataset}.
The compounds used in this dataset are extracted from the Wall Street Journal (WSJ) portion in the Penn Treebank (PTB). 

\paragraph{Marathi:} We create an annotated context-sensitive compound data for Marathi due to its unavailability. We extract compound words from Marathi grammar textbooks. For Marathi, we restrict to the same 4 semantic types as in Sanskrit (coarse setting). In our dataset, the context corresponding to 75\% data points is automatically leveraged from Wikipedia. In order to increase the difficulty level of the task, we ask one of the annotators to create the remaining 25\% data points in such a way that the same compound with a different context leads to a different semantic type. Next, we provide these compound words and context information to 3 annotators ($A, B, C$)\footnote{Note that the annotator who created the context for 25\% data points is different from these 3 annotators.} using a web-based platform. Refer to Appendix~\ref{marathi_annotation} for annotation interface. All annotators have their minimum academic qualification as Master in Arts in Marathi. These annotators have to choose the correct label from the multiple-choice options.\footnote{In case of confusion, we gave additional option to mark as {\sl ``Not sure''}.}   Finally, we use the maximum voting policy amongst annotators to get the label for each data point. 
Initially, we start with 1,000 data points for annotation. Out of 1,000, we drop 128 data points where none of the 2 annotators have an agreement.
The pair-wise inter-annotator agreement between annotators in terms of Cohen Kappa ($\kappa$) is as follows: $A-B: 0.40, B-C: 0.20$ and $A-C: 0.35$. The $\kappa \in [0.2, 0.4]$ is considered as fair agreement \cite{kappa}. 
\paragraph{Psuedo-labels for auxiliary tasks:} The benchmark datasets do not have a gold standard dependency and morphological information. We use predicted labels as pseudo-labels. For Sanskrit, we obtain pseudo-labels from the Trankit model \cite{nguyen2021trankit} trained on STBC dataset \cite{krishna-etal-2020-keep} and  morphological pseudo-labels from the LemmaTag model \cite{kondratyuk-etal-2018-lemmatag} trained on Hackathon dataset \cite{hackathon_data}.  For English, we use pseudo-labels from English XLM-R model\footnote{\url{https://spacy.io/models/en}} for morphological and dependency parsing task. For Marathi, we do not find any such data or pretrained model to obtain pseudo-labels. Therefore, we do not activate morphological tagging and dependency parsing components in the proposed system while training on Marathi data.
 \begin{table*}[bht]
\begin{small}
    \centering
\resizebox{1\textwidth}{!}{%
 \begin{tabular}{ccccccccccccccccccc}
\toprule
& &\multicolumn{8}{c}{\textbf{Sanskrit (coarse)}} &\multicolumn{8}{c}{\textbf{Sanskrit (fine-grained)}} 
 \\\cmidrule(r){3-10}\cmidrule(l){11-18}
 & &\multicolumn{4}{c}{\textbf{w/o context }} &\multicolumn{4}{c}{\textbf{w/ context}} &\multicolumn{4}{c}{\textbf{w/o context }} &\multicolumn{4}{c}{\textbf{w/ context}} 
 \\
 \cmidrule(r){2-2}\cmidrule(r){3-6}\cmidrule(l){7-10}\cmidrule(l){11-14} \cmidrule(l){15-18}
\textbf{Datasets}&\textbf{System} &\textbf{A}&\textbf{P}    & \textbf{R} &\textbf{F1}   &\textbf{A}&\textbf{P}    & \textbf{R} &\textbf{F1}&\textbf{A}&\textbf{P}    & \textbf{R} &\textbf{F1}&\textbf{A}&\textbf{P}    & \textbf{R} &\textbf{F1} 
\\\cmidrule(r){2-2}\cmidrule(r){3-6}\cmidrule(l){7-10}\cmidrule(l){11-14} \cmidrule(l){15-18}
&ISCLS19     & 77.68          & 76.00          & 71.00          & 73.00          & 77.68          & 76.00          & 71.00          & 73.00          & 70.64          & 67.53          & 63.18          & 64.58          & 70.64          & 67.53          & 63.18          & 64.58          \\
&COLING16    & 77.39          & \textbf{78.00} & 72.00          & \textbf{74.00} & 77.39          & 78.00          & 72.00          & 74.00          & -              & -              & -              & -              & -              & -              & -              & -              \\
&SanALBERT   & 72.01          & 72.10          & 70.00          & 71.40          & 77.40          & 77.31          & 73.00          & 74.20          & 69.39          & 62.58          & 58.13          & 58.22          & 75.40          & 70.14          & 61.22          & 62.04          \\
SaCTI-base&IndicALBERT & 71.87          & 51.60          & 53.90          & 52.00          & 78.63          & 77.47          & 75.83          & 76.47          & 68.06          & 53.05          & 53.19          & 52.30          & 71.93          & 57.29          & 56.49          & 55.25          \\
&mBERT       & 76.12          & 77.43          & 71.68          & 73.10          & 81.42          & 83.12          & 73.54          & 78.09          & 75.00          & 67.65          & 72.69          & 68.56          & 78.36          & 77.41          & 69.59          & 69.37          \\
&XLM-R        & 78.19          & 73.54          & 73.10          & 73.31          & 81.00          & 82.01          & 77.00          & 79.10          & 73.75          & 66.38          & 63.96          & 65.16          & 77.94          & 71.72          & 70.18          & 70.94          \\
&Ours        & \textbf{80.21} & 72.31          & \textbf{74.50} & 73.38          & \textbf{83.45} & \textbf{79.65} & \textbf{83.87} & \textbf{81.71} & \textbf{78.25} & \textbf{72.94} & \textbf{73.81} & \textbf{72.68} & \textbf{82.47} & \textbf{76.87} & \textbf{79.08} & \textbf{77.20}  \\
\cmidrule(r){2-2}\cmidrule(r){3-6}\cmidrule(l){7-10}\cmidrule(l){11-14} \cmidrule(l){15-18}
&ISCLS19     & 90.69          & 75.66          & 72.09          & 73.76          & 90.69          & 75.66          & 72.09          & 73.76          & 76.66          & 71.62          & 65.40          & 68.09          & 76.66          & 71.62          & 65.40          & 68.09          \\
&SanALBERT   & 88.17          & 66.65          & 61.31          & 62.85          & 87.84          & 64.15          & 65.44          & 63.63          & 79.38          & 69.90          & 75.34          & 67.57          & 80.73          & 71.62          & 74.60          & 72.69          \\
SaCTI-large&IndicALBERT & 87.77          & 68.82          & 49.58          & 56.05          & 92.95          & 84.98          & 74.90          & 79.32          & 79.03          & 68.00          & 64.14          & 63.67          & 83.13          & 74.23          & 79.77          & 76.11          \\
&mBERT       & 92.29          & 78.51          & 77.45          & 77.41          & 93.52          & 83.13          & 80.82          & 81.83          & 81.56          & 70.82          & 76.91          & 72.74          & 80.98          & 70.00          & 79.13          & 72.80          \\
&XLM-R        & 92.61          & 79.91          & 79.00          & 79.42          & 93.85          & 86.64          & 79.67          & 82.78          & 81.84          & 74.46          & 77.93          & 75.68          & 83.12          & 73.97          & 81.07          & 76.20          \\
&Ours        & \textbf{93.54} & \textbf{81.30} & \textbf{81.65} & \textbf{81.47} & \textbf{94.78} & \textbf{83.89} & \textbf{87.61} & \textbf{85.64} & \textbf{82.85} & \textbf{74.94} & \textbf{78.12} & \textbf{76.49} & \textbf{84.73} & \textbf{78.53} & \textbf{80.30} & \textbf{77.13}  \\
\hline
\end{tabular}}
    \caption{Evaluation on Sanskrit datasets in two levels of annotations (coarse and fine-grained) and two settings (w/o context and w/ context).  The best results are bold.  Results are averaged over 4 runs.
    The significance test between the best baseline XLM-R and the proposed system in terms of Recall/Accuracy metrics: $p < 0.01$ (as per t-test). We could not perform significance test with COLING16 (SaCTI-base-coarse-w/o) and report its results on all the datasets due to unavailability of its predictions and codebase. 
    ISCLS19 and COLING16 baselines cannot utilize the context information; therefore, we report the same numbers in w/context as w/o context.}
    \label{table:main_results}
    \end{small}
\end{table*}

 \paragraph{Hyper-parameter settings:} For the implementation of the proposed system, we modify on top of codebase by \newcite{nguyen2021trankit}. We use the following hyper-parameter settings for the best configuration of the proposed system: number of epochs as 70, batch size 50 and a embedding dropout rate of 0.3 with a learning rate of 0.001. In our multi-task loss function, we penalize dependency component's loss function by 0.01 to prioritize the performance on SaCTI. This penalty is identified based on hyper-parameter tuning. The rest of the hyper-parameters are kept the same as \newcite{nguyen2021trankit}. For multi-lingual baselines, we used Huggingface’s transformers repository \cite{wolf-etal-2020-transformers}.
 We release our codebase and datasets publicly under the licence CC-BY 4.0. All the artifacts used in this work are publicly available for the research purpose. 
 
 \paragraph{Computing infrastructure used:} We use a single GPU with Tesla P100-PCIE, 16 GB GPU memory, 3584 GPU Cores computing infrastructure for our experiments. Our proposed system takes approximately 1 hour for training SaCTI-base coarse w/o context setting dataset. 

\paragraph{Evaluation metrics:} Following \newcite{krishna-etal-2016-compound,sandhan-etal-2019-revisiting}, we report macro averaged \textbf{P}recision, \textbf{R}ecall and \textbf{F1}-score for all our experiments. 
We also report micro-averaged \textbf{A}ccuracy. We use Scikit-learn software \cite{scikit-learn} to compute these metrics.

\paragraph{Baselines:}  
We consider two context agnostic systems where \newcite[\textbf{ISCLS19}]{sandhan-etal-2019-revisiting} formulate SaCTI as a purely neural-based multi-class classification approach using static word embeddings of components of a compound and   \newcite[\textbf{COLING16}]{krishna-etal-2016-compound} deploy a hybrid system which leverages linguistically involved hand-crafted feature engineering with distributional information from Adaptor Grammar \cite{Johnson2006AdaptorGA}. The COLING16 system is the current state-of-the-art system for the SaCTI task. Next, we opt for multi-lingual contextual language models due to the lack of sufficiently large unlabelled data available for Sanskrit.
We consider three multi-lingual pretrained language models, namely, \newcite[\textbf{IndicALBERT}]{kakwani2020indicnlpsuite} which is ALBERT model trained on 12 Indic languages excluding Sanskrit, BERT \cite[\textbf{mBERT}]{devlin-etal-2019-bert} trained on 104 languages having largest Wikipedia's excluding Sanskrit and \newcite[\textbf{XLM-R}]{conneau-etal-2020-unsupervised} trained on 100 languages including Sanskrit. Finally, we consider a mono-lingual ALBERT model~\cite[\textbf{SanALBERT}]{jivnesh_eval} trained on Sanskrit corpus \cite{hellwig2010dcs} from scratch. In all the contextual baselines, we pass the classification token [CLS] representation of a sentence pair (compound word and its context separated by [SEP] token) to the classification head.
\textbf{Ours:}  This is our proposed system from \S~\ref{proposed-system}.

\subsection{Results}
\label{main-results}
Table~\ref{table:main_results} shows the performance for the best performing configurations of all the baselines on the test set of benchmark datasets for SaCTI.
We report results on two levels of annotations (coarse and fine-grained) and two settings (w/o context and w/ context).\footnote{For the systems that require context, we feed the compound word only as the context.}
Except for ISCLS19 and COLING16 systems, all systems utilize available context along with components of a compound.\footnote{These baselines cannot utilize the context; therefore, we report the same numbers in w/context as w/o context.}
XLM-R reports the best performance among all the baselines while using the context information. 

Our proposed system outperforms all the competing systems in terms of all the evaluation metrics and reports $6.1$ points (A) and $7.7$ points (F1) absolute gain with respect to the current state-of-the-art system COLING16 (on SaCTI-base coarse w/context dataset). COLING16 outperforms the proposed system  on SaCTI-base coarse w/o context. 
Notably, our proposed system outperforms the strong baseline XLM-R with large margins in fine-grained setting on SaCTI-base dataset (low-resourced setting). This confirms the usefulness of the proposed system in low-resourced settings with fine-grain labels.
The large performance gap between the proposed system with context and COLING16/ISCLS19 baselines illustrates the efficacy of using contextual information and syntactic information. 
Summarily, we mark new state-of-the-art results with the help of the novel architecture, where the contextual component is integrated with syntax-based auxiliary tasks such as morphological tagging and dependency parsing.
We find a similar trend in performance for the SaCTI-large dataset.

\section{Analysis}
\label{analysis}
In this section, we dive deep into the proposed system architecture for a detailed analysis as well as generalizability. We use SaCTI-base coarse dataset in the w/ context setting for the analysis.
\paragraph{(1) Ablation analysis:} Here, we investigate the contribution of various components towards the overall improvements of the proposed system. Table~\ref{table:ablation} reports ablations in terms of all the evaluation metrics when a particular component is inactivated from the proposed system. For example, ``-DP'' denotes the system where the dependency parsing component is removed from the proposed system. We see that elimination of any of the components deteriorates the performance. Table~\ref{table:ablation} illustrates that `context' component is the most critical towards improvements. Also, the deletion of the `BiAFF' component has the second largest impact on the final performance.\footnote{In the absence of the proposed `BiAFF' component, we use [CLS] token for the sentence-level prediction, where this system is similar to XLM-R + DP + morph.} 

\begin{table}[h]
\centering
\begin{adjustbox}{width=0.45\textwidth}
\small
\begin{tabular}{|c|c|c|c|c|}
\hline
\textbf{System}                           & \textbf{A}     & \textbf{P}     & \textbf{R}  & \textbf{F1}    \\ \hline
Ours        & 83.45 & 79.65 & 83.87 & 81.71 \\\hline
-context    & 80.21 & 72.31 & 74.50 & 73.38 \\\hline
-BiAFF      & 81.00 & 82.01 & 77.00 & 79.10 \\\hline
-morph  & 82.87 & 80.00 & 81.21 & 80.60 \\\hline
-DP    & 81.89 & 79.35 & 81.62 & 80.26 \\ \hline
-morph -DP & 81.50 & 82.34 & 77.67 & 79.93 \\ \hline
\end{tabular}
\end{adjustbox}
\caption{Ablations of the proposed system
in terms of all the metrics. Each ablation deletes a single component from the proposed system. For example, “-DP” deletes the dependency parsing task from the proposed system.} 
\label{table:ablation}
\end{table}

\begin{figure}[!h]
\centering
\subfloat[]{\includegraphics[width=1.5in]{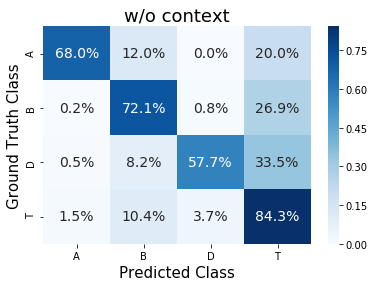}}
\subfloat[]{\includegraphics[width=1.5in]{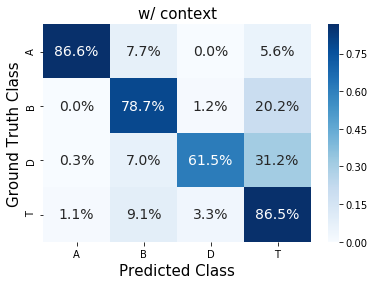}}\\
\caption{The confusion matrix for the proposed system trained (a) w/o context (b) w/ context. Semantic types: \textit{\textbf{A}vyay\={\i}bh\={a}va}, \textit{\textbf{B}ahuvr\={\i}hi}, \textit{\textbf{D}vandva}, and \textit{\textbf{T}atpuru\d{s}a}} 
\label{fig:error} 
\end{figure}

 \begin{table*}[bht]
\begin{small}
    \centering
\resizebox{1\textwidth}{!}{%
 \begin{tabular}{cccccccccccccccccc}
\toprule
 &\multicolumn{8}{c}{\textbf{English}} &\multicolumn{8}{c}{\textbf{Marathi}} 
 \\\cmidrule(r){2-9}\cmidrule(l){10-17}
  &\multicolumn{4}{c}{\textbf{w/o context}} &\multicolumn{4}{c}{\textbf{w/ context}} &\multicolumn{4}{c}{\textbf{w/o context}} &\multicolumn{4}{c}{\textbf{w/ context}}  \\
 \cmidrule(r){2-5}\cmidrule(l){6-9}\cmidrule(l){10-13} \cmidrule(l){14-17}
\textbf{System} & \textbf{A}     & \textbf{P}     & \textbf{R}     & \textbf{F1}    & \textbf{A}     & \textbf{P}     & \textbf{R}     & \textbf{F1}    & \textbf{A} & \textbf{P} & \textbf{R} & \textbf{F1} & \textbf{A} & \textbf{P} & \textbf{R} & \textbf{F1} \\
 \cmidrule(r){2-5}\cmidrule(l){6-9}\cmidrule(l){10-13} \cmidrule(l){14-17}
ISCLS19     & 67.43          & 71.81          & 64.38          & 66.26          & 67.43          & 71.81          & 64.38          & 66.26          & 68.62          & 70.78          & 52.89          & 56.85          & 68.62          & 70.78          & 52.89          & 56.85          \\
IndicALBERT & 68.23          & 57.25          & 59.70          & 57.96          & 70.22          & 68.01          & 69.31          & 68.65          & 67.65          & 33.33          & 45.15          & 38.08          & 60.00          & 45.17          & 40.62          & 40.95          \\
mBERT       & 70.98          & 70.00          & 69.87          & 69.97          & 72.48          & 76.08          & 73.47          & 74.30          & 77.45          & 58.70          & 61.28          & 59.48          & 71.05          & 51.45          & 53.34          & 52.18          \\
BERT        & 74.19          & 72.79          & 71.12          & 71.62          & 74.71          & 75.83          & 74.48          & 75.12          & 78.43          & 71.21          & 68.04          & 69.20          & 75.43          & 67.07          & 66.07          & 66.43          \\
XLM-R        & 72.21          & 71.53          & 68.88          & 69.20          & 74.33          & 77.12          & 76.86          & 76.57          & 76.47          & 67.36          & 62.90          & 63.33          & 74.56          & 65.50          & 62.07          & 62.75          \\
Ours        & \textbf{74.69} & \textbf{74.79} & \textbf{75.19} & \textbf{75.19} & \textbf{77.81} & \textbf{79.17} & \textbf{79.19} & \textbf{79.12} & \textbf{78.12} & \textbf{71.98} & \textbf{70.00} & \textbf{70.57} & \textbf{80.43} & \textbf{66.54} & \textbf{77.00} & \textbf{69.12}     \\
\hline
\end{tabular}}
    \caption{Evaluation on English and Marathi languages. The best results are bold. The significance test between the best baseline XLM-R and our system in terms of Recall/Accuracy metrics: $p < 0.01$ (as per t-test).  ISCLS19 do not have power to utilize the context information; therefore, we report the same numbers in w/context as w/o context.}
    \label{table:multilingual_results}
    \end{small}
\end{table*} 
\paragraph{(2) How effective is the proposed system in reducing confusion between conflicting classes?}
Figure~\ref{fig:error} illustrates the confusion matrices in w/ context and w/o context scenarios. We observe a similar trend in both the scenarios.  (1) Both systems mis-classify the predictions into the most populated type (\textit{Tatpuru\d{s}a}). This can be attributed to the imbalanced nature of the dataset.\footnote{In this work, we do not consider any strategy to tackle imbalanced classification. We plan to address this in future.} (2) The confusion between \textit{Avyay\={\i}bh\={a}va} and \textit{Tatpuru\d{s}a} is due to these compounds having their first component as an indeclinable word. Notably, the system with context is able to reduce confusion by 15\%. (3) One of the reasons for conflict between \textit{Tatpuru\d{s}a} and \textit{Bahuvr\={\i}hi} is due to the specific subcategory of both the classes where the first component is a negation. With the help of enriched information, a system with context can reduce this miss-classification by 7\%. Summarily, the system with contextual information always performs superior to one with no context.
This substantiates the importance of contextual information in the proposed system. 

\paragraph{(3) How well can we generalize the proposed system for other languages?}
The primary motivation is to illustrate the efficacy of our language agnostic approach. The semantic type of compounds of the language of interest need not be similar to that of Sanskrit for the model to work. It is purely language agnostic model. To study the generalization ability of the proposed system, we consider 2 additional languages, namely,  
English (en) and Marathi (ma).  We choose English due to its availability of context-sensitive annotated data and Marathi due to its closeness to Sanskrit. We freshly created annotated task-specific context-sensitive data for Marathi (\S~\ref{datasets}) as no such dataset was previously available. Table~\ref{table:multilingual_results} reports the results for these two languages. For English, all the baselines (with context) improve over their counterpart (without context). However, we do not find similar trend in Marathi possibly due to (1) lack of sufficient task-specific data, and (2) lack of both the auxiliary task~\footnote{We could not activate auxiliary tasks due to lack of datasets for Marathi.}.
For both languages, our system consistently outperforms all the competing systems.  Across both the languages, it shows the average absolute gain of $4.7$ points (A) and $4.4$ points (F1) compared to the strong baseline XLM-R.  Summarily, these empirical results prove the proposed approach's efficacy in languages other than Sanskrit.

\paragraph{(4) Multi-lingual training and zero-shot cross-lingual transfer experiments for Marathi:} Here, we investigate the transferability of the SaCTI task for low-resourced languages. We experiment with the Marathi language (w/ context). Since the label space of Marathi is the same as that of the SaCTI coarse dataset, this makes it possible to experiment with cross-lingual zero-shot transfer and multi-lingual training. There is an isomorphic semantic type system with 4 types for Marathi as is the case for most of the Indian languages, due to a close connection with / inheritance from Sanskrit.  In Table~\ref{table:multi-lingual_experiments}, we consider the mono-lingual (training on Marathi) results as a baseline.  In multi-lingual training, we train the proposed system with mix of Marathi and SaCTI-base coarse dataset and evaluate on test set of Marathi.\footnote{Here, we use pretrained models of Sanskrit for both the auxiliary tasks to obtain psuedo-labels for Marathi.} In the zero-shot transfer, we leverage the model trained on the SaCTI-base coarse dataset to get predictions on the test set of Marathi. In multi-lingual training experiment, we observe substantial improvements ($3.4$ points F1) over mono-lingual training. However, zero-shot cross-lingual transfer does not show encouraging results, where system predicts \textit{Tatpuru\d{s}a} type for majority test samples ($80\%$ predictions).
\begin{table}[h]
\centering
\begin{adjustbox}{width=0.45\textwidth}
\small
\begin{tabular}{|c|c|c|c|c|}
\hline
\textbf{Tasks}                           & \textbf{A}     & \textbf{P}     & \textbf{R}  & \textbf{F1}    \\ \hline
ours (Marathi)         & 80.43          & 66.54          & 77.00          & 69.12          \\\hline
zero-shot transfer     & 48.09          & 33.85          & 48.36          & 31.08          \\\hline
multi-lingual training & \textbf{83.10} & \textbf{71.00} & \textbf{80.48} & \textbf{73.70}\\\hline
\end{tabular}
\end{adjustbox}
\caption{Performance of the multilingual training and cross-lingual zero-shot transfer on Marathi.} 
\label{table:multi-lingual_experiments}
\end{table}
\begin{figure*}[!h]
\centering
\subfloat[]{\includegraphics[width=1.5in]{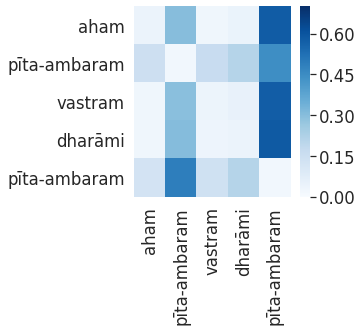}}
\subfloat[]{\includegraphics[width=1.5in]{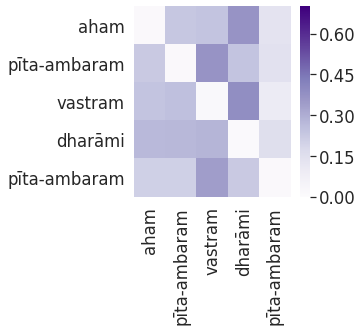}}
\subfloat[]{\includegraphics[width=1.5in]{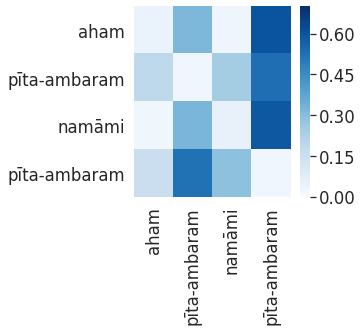}}
\subfloat[]{\includegraphics[width=1.5in]{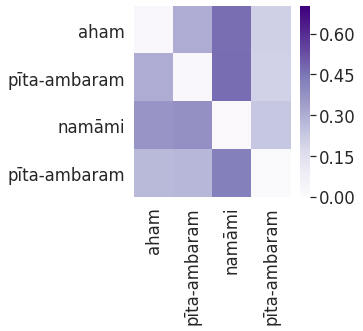}}\\
\caption{Attention heatmaps for the SaCTI (Blue) and dependency parsing (Purple) tasks.  The SaCTI heatmap shows how different context words contribute towards final prediction (Blue) and dependency parsing heatmap (Purple) serve as proxy for interpretation.  We illustrates how the same compound ({\sl p\={\i}ta-ambaram}) in two different contexts [(a-b) {\sl aham \textbf{p\={\i}ta-ambaram} vastram dhar\={a}mi} (I wear a yellow cloth) and (c-d) {\sl aham \textbf{p\={\i}ta-ambar\={a}m} nam\={a}mi} (I pray to the Lord Vi\d{s}nu)] leads to different semantic type predictions (\textit{Tatpuru\d{s}a}: yellow cloth and \textit{Bahuvr\={\i}hi}: Lord Vi\d{s}nu). In the dependency heatmap of the first case, \textit{p\={\i}ta-ambaram} focuses on \textit{vastram} (cloth) and in latter case it focuses on \textit{nam\={a}mi} (the action of praying).}
\label{fig:probing} 
\end{figure*}
\paragraph{(5) Probing analysis:}
Here, we probe the attention modules of the proposed system to investigate (1) How do different context words contribute towards final prediction? (2) Do these attentions serve as a proxy for explainability of correct/incorrect predictions? Figure~\ref{fig:probing} illustrates attention heatmaps for the SaCTI (Blue) and dependency parsing (Purple) tasks.  The SaCTI heatmap shows how different context words contribute towards final prediction (Blue) and dependency parsing heatmap (Purple) serves as proxy for interpretation. We notice in the SaCTI attentions that all words mostly focus on the target compound word.  Figure~\ref{fig:probing} illustrate how the same compound ({\sl p\={\i}ta-ambaram}) in two different contexts [(a-b) {\sl aham \textbf{p\={\i}ta-ambaram} vastram dhar\={a}mi} (I wear a yellow cloth) and (c-d) {\sl aham \textbf{p\={\i}ta-ambar\={a}m} nam\={a}mi} (I pray to the Lord Vi\d{s}nu)] leads to different semantic type predictions (\textit{Tatpuru\d{s}a}: yellow cloth and \textit{Bahuvr\={\i}hi}: Lord Vi\d{s}nu). In the dependency heatmap of the first case, \textit{p\={\i}ta-ambaram} focuses on \textit{vastram} (cloth) and in latter case, it focuses on \textit{nam\={a}mi} (the action of praying).  As per the grammatical rules, the morphological tagging task correctly predicts the gender information as neuter and masculine in these cases, respectively.
Thus, this probing analysis suggests that auxiliary tasks not only help add complementary signals to the system but also serve as a proxy for explainability. 
\paragraph{(6) Additional auxiliary tasks:}
\label{additional_tasks}
  With our proposed multi-task learning approach, we experiment with a few more additional sequence labeling auxiliary tasks (on SaCTI-base w/ context dev set), namely, the prediction of the case grammatical category (C), lemma prediction (L) and prediction of a relation (R) between modifier and its headword. The results in Table~\ref{table:additional_aux} show that except for the relation prediction task, all the remaining auxiliary tasks report improvements over the base system (with no auxiliary task). However, none of the combinations of these auxiliary tasks could outperform the proposed combination of morphological parsing and dependency parsing tasks. Therefore, we do not consider these additional auxiliary tasks in our final system.
\begin{table}[!h]
\centering
\begin{adjustbox}{width=0.45\textwidth}
\small
\begin{tabular}{|c|c|c|c|c|}
\hline
\textbf{Tasks}                           & \textbf{A}     & \textbf{P}     & \textbf{R}  & \textbf{F1}    \\ \hline
BiAFF            & 87.99 & 85.48 & 87.90  & 86.64 \\\hline
+case (C)            & 87.64 & 85.65 & 88.90  & 87.19 \\\hline
+morph (M)             & 88.01 & 88.90  & 85.75 & 87.26 \\\hline
+relation (R)     & 86.61 & 85.53 & 84.99 & 85.22 \\\hline
+lemma (L)            & 87.43 & 88.27 & 86.21 & 87.18 \\\hline
+Dep. parse (DP)               & 89.14 & 86.49 & 90.10  & 88.01 \\\hline \hline
M+C       & 87.55 & 88.30  & 85.28 & 86.72 \\\hline
M+C+L & 87.08 & 87.74 & 84.98 & 86.30  \\\hline
M+C+R & 86.10  & 86.04 & 83.28 & 84.55 \\\hline
M+DP         & \textbf{88.11} & \textbf{86.12} & \textbf{89.23} & \textbf{88.43} \\ \hline
\end{tabular}
\end{adjustbox}
\caption{The comparison (on SaCTI-base w/ context dev set) in between auxiliary tasks. `+' denotes a system where the corresponding task is integrated with BiAFF.} 
\label{table:additional_aux}
\end{table}
\paragraph{(7) Web-based tool:}
 We deploy our pretrained models as a web-based tool which facilitates the following advantages: (1) A naive user with no prior deep-learning expertise can use it for pedagogical purposes. (2) It can serve as a semi-supervised annotation tool keeping a human in the loop for the error corrections.  (3) Our tool helps the user interpret the model prediction using model confidence on each semantic type and the probing analysis. Refer to Appendix~\ref{web-based-tool} for our web-based tool's interface. (4) It can be used for any general purpose classification task.
\section{Related work}
\label{related_section}
\paragraph{English Noun Compound Interpretation} 
Prior to the deep learning era, various machine learning-based approaches have been proposed for Noun Compound Identification \cite{kim-baldwin-2005-automatic,o-seaghdha-copestake-2009-using,tratz-hovy-2010-taxonomy}. With the advent of deep-learning based approaches, \newcite{dima-hinrichs-2015-automatic} and \newcite{fares-etal-2018-transfer} proposed a neural-based architecture where concatenated representations of a compound were fed to a feed-forward network to predict a semantic relation between the compound's components. \newcite{shwartz-waterson-2018-olive} proposed an approach that combines labeling with paraphrasing. 
Recently, \newcite{ponkiya-etal-2021-framenet} proposed a novel approach using semantic label repository \cite[FrameNet]{ponkiya-etal-2018-towards} where continuous label space embeddings are used to predict unseen labels. To the best of our knowledge, a context has never been used for the classification task for NCI. In paraphrasing line of modeling, \newcite{ponkiya-etal-2020-looking} formulates paraphrasing as ``fill-in-the-blank'' problem to predict the ``missing'' predicate or preposition using pretrained language models.

\paragraph{Sanskrit Compound Type Identification} task has garnered considerable attention of the researchers in the last decade.
In order to decode the meaning of a Sanskrit compound, it is essential to figure out its constituents \cite{huet2010,mittal-2010-automatic,hellwig-nehrdich-2018-sanskrit}, how the constituents are grouped \cite{Kulkarni2011StatisticalCP}, identify the semantic relation between them \cite{anil_thesis} and finally generate the paraphrase of the compound \cite{Kumar2009SanskritCP}.
 \newcite{Satuluri2013GenerationOS} and \newcite{kulkarni2013} proposed a rule-based approach where around 400 rules mentioned in P\={a}\d{n}ini's grammar \cite{panini} were analysed from the perspective of compound generation and type identification, respectively.
 Recently, \newcite{sandhan-etal-2019-revisiting} investigated whether a purely engineering data-driven approach competes with the performance of a linguistically motivated hybrid approach by \newcite{krishna-etal-2016-compound}. 
 Summarily, no attention has been given to incorporating contextual information, which is crucial and cheaply available. We address this research gap and mark the new state-of-the-art results with substantial improvements.

\section{Conclusion and Discussion}
 This work focused on Sanskrit compound type identification, where the task is to decode the semantic information hidden in the compound, which can be {\sl context-sensitive}. This poses a limitation to the existing \textit{context agnostic} approaches, thus we propose a novel multi-task learning architecture which incorporates the contextual information and also enriches it with complementary syntactic information using morphological tagging and dependency parsing auxiliary tasks. Our probing analysis showcased that these auxiliary tasks also serve as a proxy for model prediction explainability. To the best of our knowledge, this is the first time that the importance of these auxiliary tasks has been showcased for SaCTI.
Our experiments on benchmark datasets showed that the proposed system provides stunning improvements with $6.1$ points (A) and $7.7$ points (F1) absolute gain compared with the current state-of-the-art system. Our fine-grained analysis showcased some light on the inner engineering of the proposed system. 
Our multi-lingual experiments on English and Marathi languages proved the efficacy of the proposed system in other languages.

 We limit our study to the {\sl purely engineering} data-driven settings.
 We plan to extend the current work by augmenting logical rules \cite{li-srikumar-2019-augmenting,nandwani2019primal} derived from P\={a}\d{n}inian grammar in the proposed approach. 

\paragraph{Ethics Statement:}
We do not foresee any ethical concerns with the
work presented in this manuscript.

\section*{Acknowledgements}
We thank Amba Kulkarni for providing annotations for SaCTI datasets.
We are grateful to Pavankumar Satuluri for helping us with insightful discussion. We would like to thank Narein Rao, Rathin Singha and the anonymous reviewers for their constructive feedback towards improving this work.
We thank Om Adideva Paranjay and Ayush Daksh for helping us with the data pre-processing step.  We would like to thank Jayesh Lakade, Prachi Rahangade and Aakash Biswas for helping us with the Marathi compound raw data collection.  We thank Varsha Sandhan, Chitra Manjarekar, Khyati Deshpande and Rachit Bodhare for helping us with the Marathi data annotations.  The work of the first author is supported by the TCS Fellowship under the Project TCS/EE/2011191P. 

\bibliography{anthology,custom}
\bibliographystyle{acl_natbib}
\appendix
\begin{figure*}[!hbt]
\centering
\includegraphics[width=0.8\textwidth]{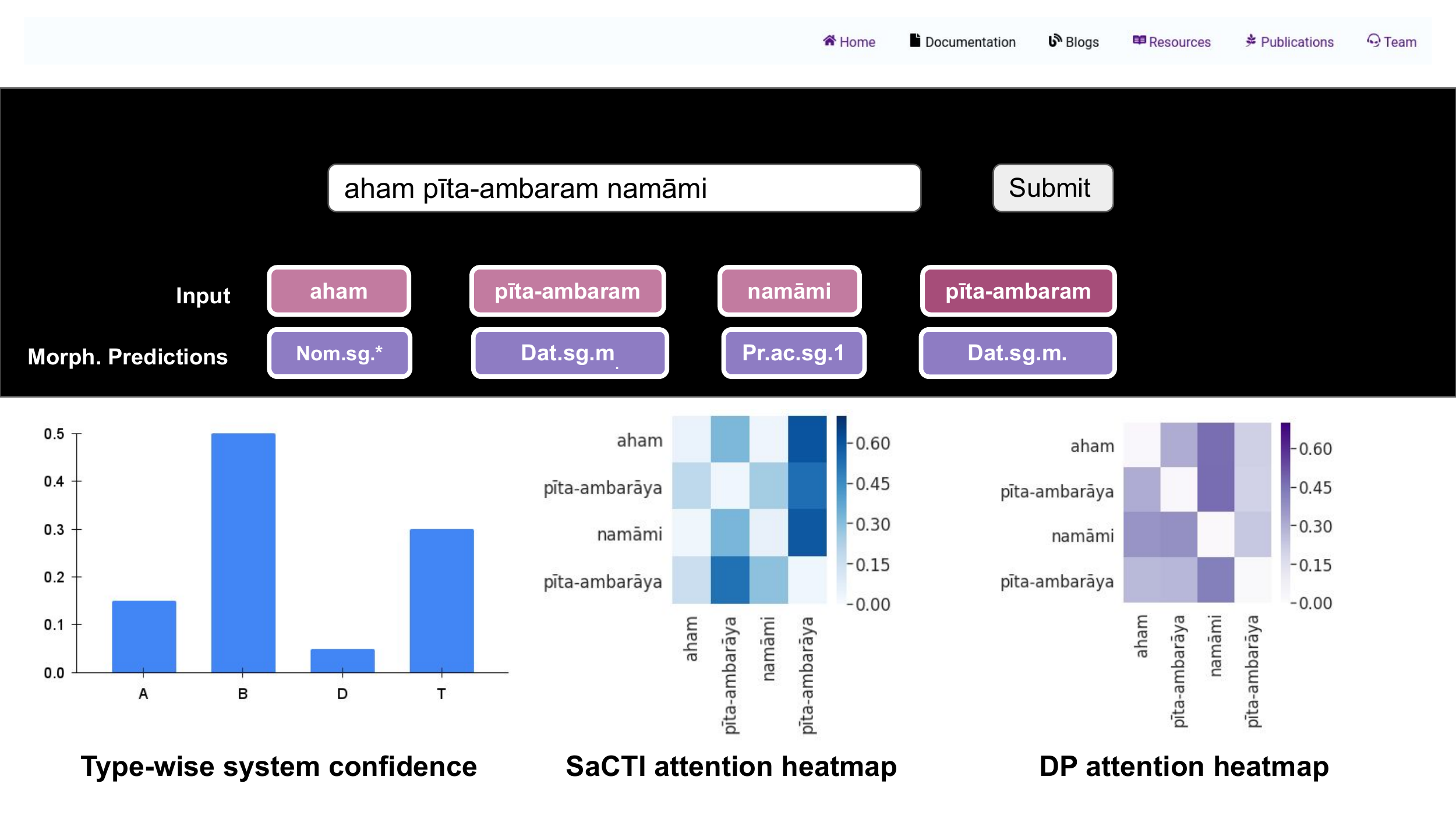}
\caption{Illustration of web-based tool integrated with our best performing pretrained system. Input: ``{\sl aham p\={\i}ta-ambaram nam\={a}mi}'' (Translation: ``I pray to P\={\i}t\={a}mbara (Lord Vishnu).'') where `{\sl p\={\i}ta-ambaram}' is a compound word. Our interface shows predicted morphological tags (color-coded with violet boxes), type-wise system confidence (bar plot), attention heatmaps.} 
\label{fig:interface_interpret} 
\end{figure*}


\section{Web-based tool}
\label{web-based-tool}
 We deploy our pretrained models as a web-based tool. Figure~\ref{fig:interface_interpret} illustrates the web-based tool integrated with our best performing system. Input: ``{\sl aham p\={\i}ta-ambaram nam\={a}mi}'' (Translation: ``I pray to P\={\i}t\={a}mbara (Lord Vishnu).'') where `{\sl p\={\i}ta-ambaram}' is a compound word. Our interface shows predicted morphological tags (color-coded with violet boxes), type-wise system confidence (bar plot), attention heatmaps. Our tool helps the user interpret the model prediction using model confidence on each semantic type.
 \begin{figure*}[!h]
\centering
\subfloat[]{\includegraphics[width=1.5in,height=2in]{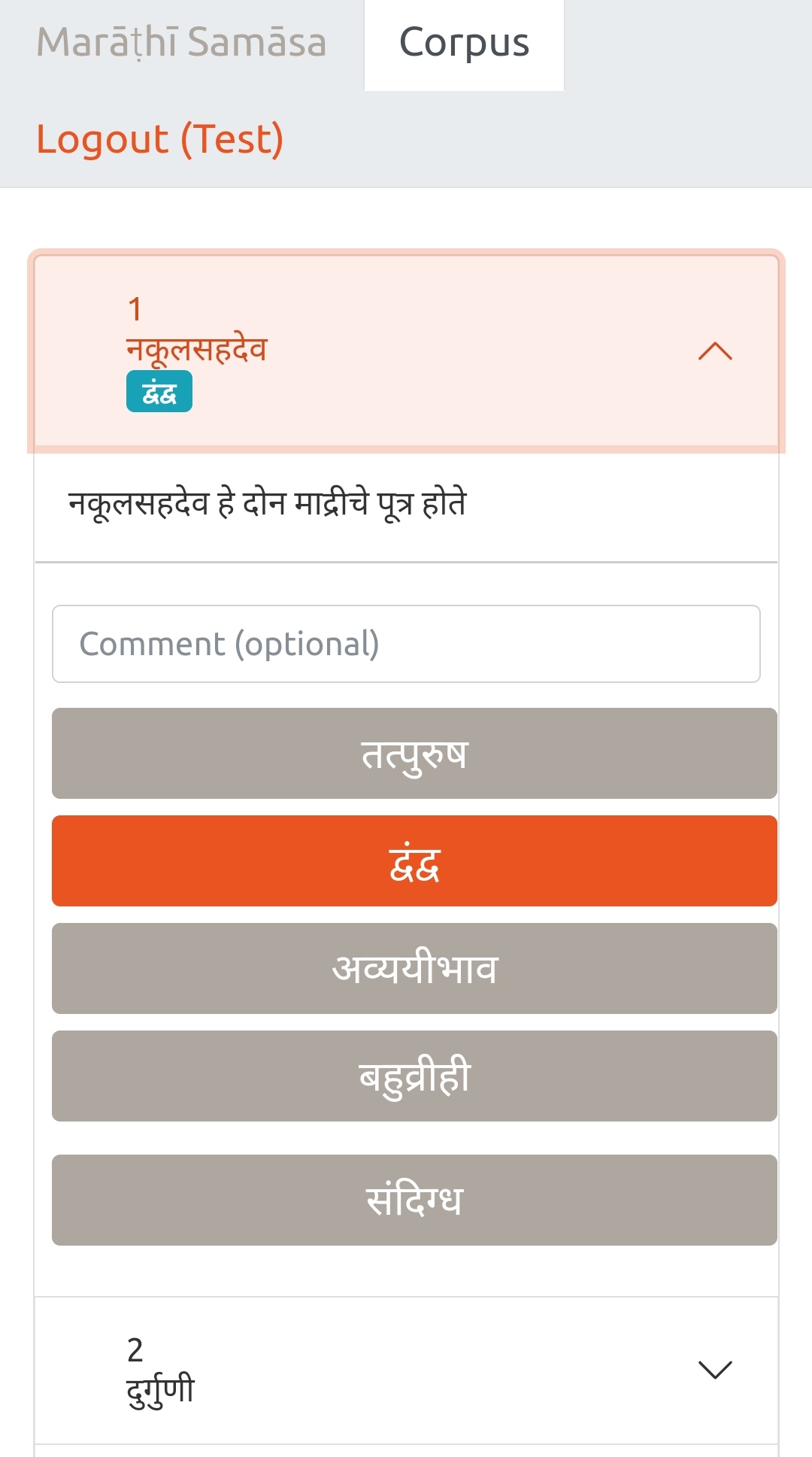}}\qquad \qquad
\subfloat[]{\includegraphics[height=1.5in,height=2in]{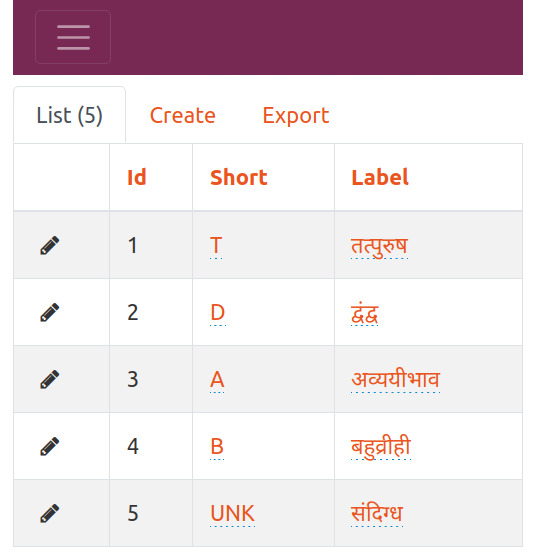}}
\caption{Illustration of our web-based mobile friendly (a) annotation interface, where the task is to select the correct option from multiple-choices for the given compound with the context. (b) Administrative interface, where the administrators can easily control the class labels and classification context as well as export annotations.} 
\label{fig:annotation} 
\end{figure*}
\section{Marathi annotation details}
\label{marathi_annotation}
We have built a user-friendly annotation tool for general-purpose classification tasks. The tool is a Flask-based \cite{ronacher2011flask}  user-friendly web application styled by Bootstrap 5 \cite{bootstrap51}, and sports a simple administrative interface that lets the administrators easily control the class labels and classification context as well as export annotations.
Figure~\ref{fig:annotation} shows our web-based mobile friendly annotation interface, where the task is to select the correct option from multiple-choices for the given compound with the context.  There are 4 broad categories of semantic types in Marathi. In case of confusion, we gave additional option to mark as {\sl ``Not sure''}. We also provide an option to add a comment to convey additional information about ambiguity or the concern for the corresponding example. 
\end{document}